\documentclass{article}

\usepackage[preprint]{neurips_2025}

\usepackage[utf8]{inputenc} 
\usepackage[T1]{fontenc}    
\usepackage{hyperref}       
\usepackage{url}            
\usepackage{booktabs}       
\usepackage{amsfonts}       
\usepackage{nicefrac}       
\usepackage{microtype}      
\usepackage{xcolor}         
\usepackage{pifont}
\definecolor{mygray}{gray}{.9}
\definecolor{ForestGreen}{RGB}{34,139,34}
\usepackage{comment} 
\usepackage{arabtex,utf8}
\newcommand{\cmark}{\textcolor{green}{\ding{51}}} 
\newcommand{\xmark}{\textcolor{red}{\ding{55}}}   

\usepackage{multirow}
\usepackage{colortbl}
\usepackage{float}
\usepackage{algpseudocode}
\usepackage[ruled]{algorithm2e}
\usepackage{caption}
\usepackage{subcaption}
\usepackage{enumitem}
\usepackage{etoolbox}
\usepackage[normalem]{ulem}
\usepackage{dblfloatfix}
\usepackage{wrapfig}        
\usepackage{tcolorbox}

\usepackage{subcaption}    
\usepackage{tabularx}          
\usepackage{array}  
\usepackage{graphicx}
\definecolor{mygray}{gray}{0.9}
\usepackage{amsmath}
\usepackage{adjustbox}

\title{Tell me Habibi, is it Real or Fake? \adjustbox{valign=c}{\includegraphics[height=1.0cm]{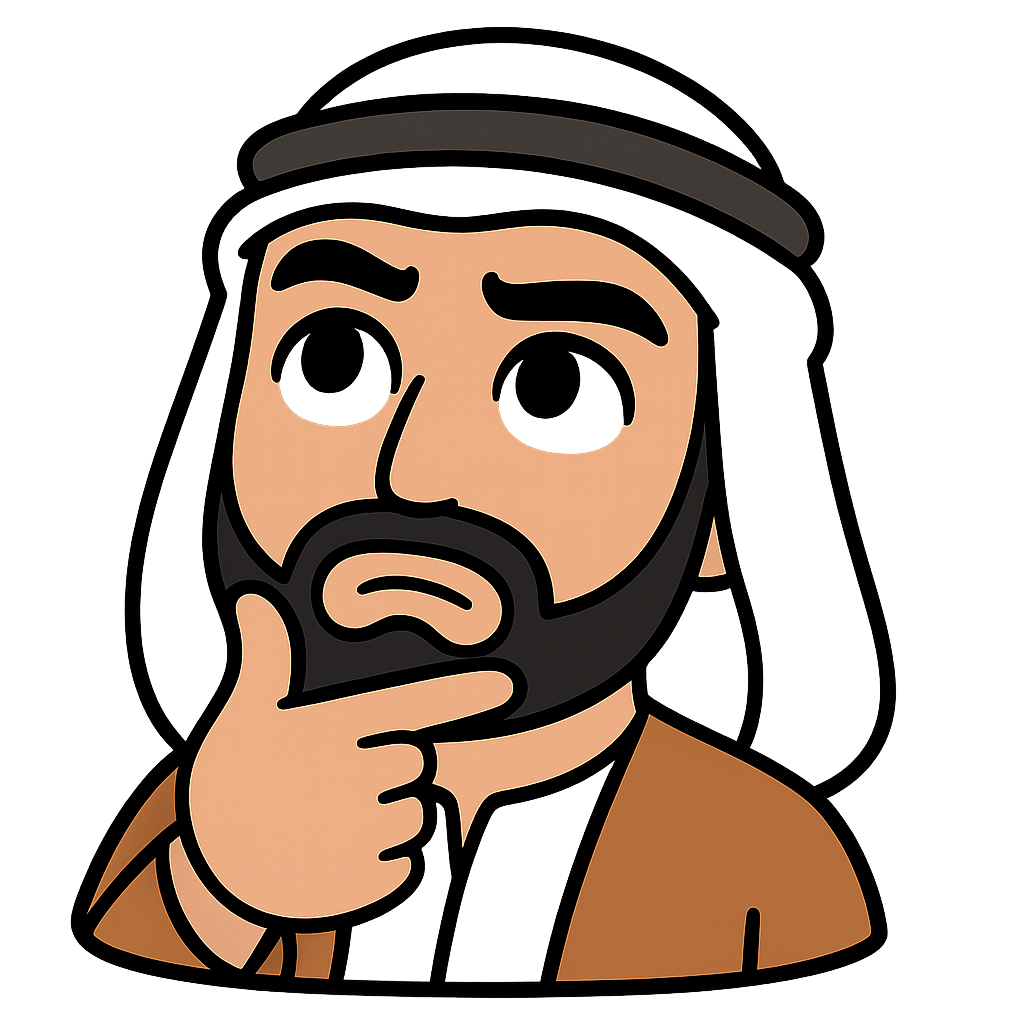}}}

\author{Kartik Kuckreja,$^\lambda$ Parul Gupta,$^\xi$ Injy Hamed,$^\lambda$ \\\textbf{Thamar Solorio,$^\lambda$ Muhammad Haris Khan,$^\lambda$ Abhinav Dhall$^\xi$}\\
$^\lambda$MBZUAI \quad 
$^\xi$Monash University\\
  {\tt \{kartik.kuckreja,injy.hamed,thamar.solorio,muhammad.haris\}@mbzuai.ac.ae} \\
  {\tt \{parul,abhinav.dhall\}@monash.edu}\\ 
  }
\begin{document}

\maketitle
\begin{myabstract}
Deepfake generation methods are evolving fast, making fake media harder to detect and raising serious societal concerns. Most deepfake detection and dataset creation research focuses on monolingual content, often overlooking the challenges of multilingual and code-switched speech, where multiple languages are mixed within the same discourse. Code-switching, especially between Arabic and English, is common in the Arab world and is widely used in digital communication. This linguistic mixing poses extra challenges for deepfake detection, as it can confuse models trained mostly on monolingual data. To address this, we introduce \textbf{ArEnAV}, the first large-scale Arabic-English audio-visual deepfake dataset featuring intra-utterance code-switching, dialectal variation, and monolingual Arabic content. It \textbf{contains 387k videos and over 765 hours of real and fake videos}. Our dataset is generated using a novel pipeline integrating four Text-To-Speech and two lip-sync models, enabling comprehensive analysis of multilingual multimodal deepfake detection. We benchmark our dataset against existing monolingual and multilingual datasets, state-of-the-art deepfake detection models, and a human evaluation, highlighting its potential to advance deepfake research. The dataset can be accessed \href{https://huggingface.co/datasets/kartik060702/ArEnAV-Full}{here}.
\end{myabstract}

\setcode{utf8}

\section{Introduction}

Deepfake technologies, involving the artificial generation and manipulation of audio-visual content, have rapidly advanced, significantly complicating the task of distinguishing real media from synthetic creations. The potential misuse of deepfakes for misinformation, defamation, or impersonation presents profound societal risks, driving substantial research into their detection. Although initial deepfake research primarily focused on manipulating individual modalities, audio-only (\citet{todiscoASVspoof2019}) or video-only (\citet{jiangDeeperForensics12020,kwonKoDF2021,liCelebDF2020}), recent developments increasingly consider joint manipulation of audio and visual streams for more realistic synthesis.

A significant gap remains in existing deepfake datasets (Table \ref{tab:existing_datasets}), which largely overlook multilingual scenarios, particularly code-switching (CSW), despite its global prevalence among bilingual speakers. 
In the Arab world, CSW is a prominent feature of daily communication, serving not only as a linguistic tool but also as a marker of cultural identity and social context (\cite{hamed2025survey}). 
Arabic speakers frequently alternate between Arabic and English within the same sentence, such as: 
\<مهم جداً> \textit{deepfake detection} \<موضوع ال>
(The topic of \textit{deepfake detection} is very important). 
Moreover, with Arabic being a diglossic language, speakers also switch between Modern Standard Arabic (MSA) and regional dialects (\cite{mubarak2021qasr}). 
Recent studies provide compelling evidence of how common CSW is in Arabic-English contexts. The ZAEBUC-Spoken corpus (\citet{hamed2024zaebuc}) reveals that approximately 19\% of all non-annotation-only utterances exhibit CSW, having an average of 44\% English words. 
The corpus also highlights the presence of CSW between Arabic variants. 
Similarly, the ArzEn corpus (\citet{hamed-etal-2020-arzen})  
demonstrates a high frequency of CSW, 
where 63\% of utterances involve CSW with approximately 19\% of words being English. 
These findings highlight the extent to which Arabic-English code-switching is not merely an incidental phenomenon, but a widespread communicative strategy that deeply influences spoken discourse. However, despite its ubiquity, deepfake detection systems remain largely ill-equipped to handle such language alternation, focusing predominantly on monolingual data. Addressing this oversight, our work seeks to bridge this critical gap by introducing the first Arabic-English CSW audio-visual deepfake dataset, thus advancing the field toward more relevant detection systems. Our core contributions are as follows:

\begin{itemize}
\item Introduction of ArEnAV, the first large-scale Arabic-English audio-visual deepfake dataset featuring intra-utterance code-switching and dialectal variation, including both bilingual and diglossic switching across Modern Standard Arabic, Egyptian, Levantine, and Gulf dialects, addressing a critical gap in multilingual deepfake research.

\item A novel data generation pipeline tailored to English and Arabic (MSA and dialect-rich content), integrating four TTS (Text to Speech) and two lip-sync models.
\item A comprehensive analysis contrasting our dataset against existing monolingual and multilingual datasets,  existing state-of-the-art (SOTA) deepfake detection models, and a detailed User Study, underscoring its unique difficulty in detection by models and humans alike. 
\end{itemize}

\begin{figure*}[t]
  \centering
  \includegraphics[width=\textwidth]{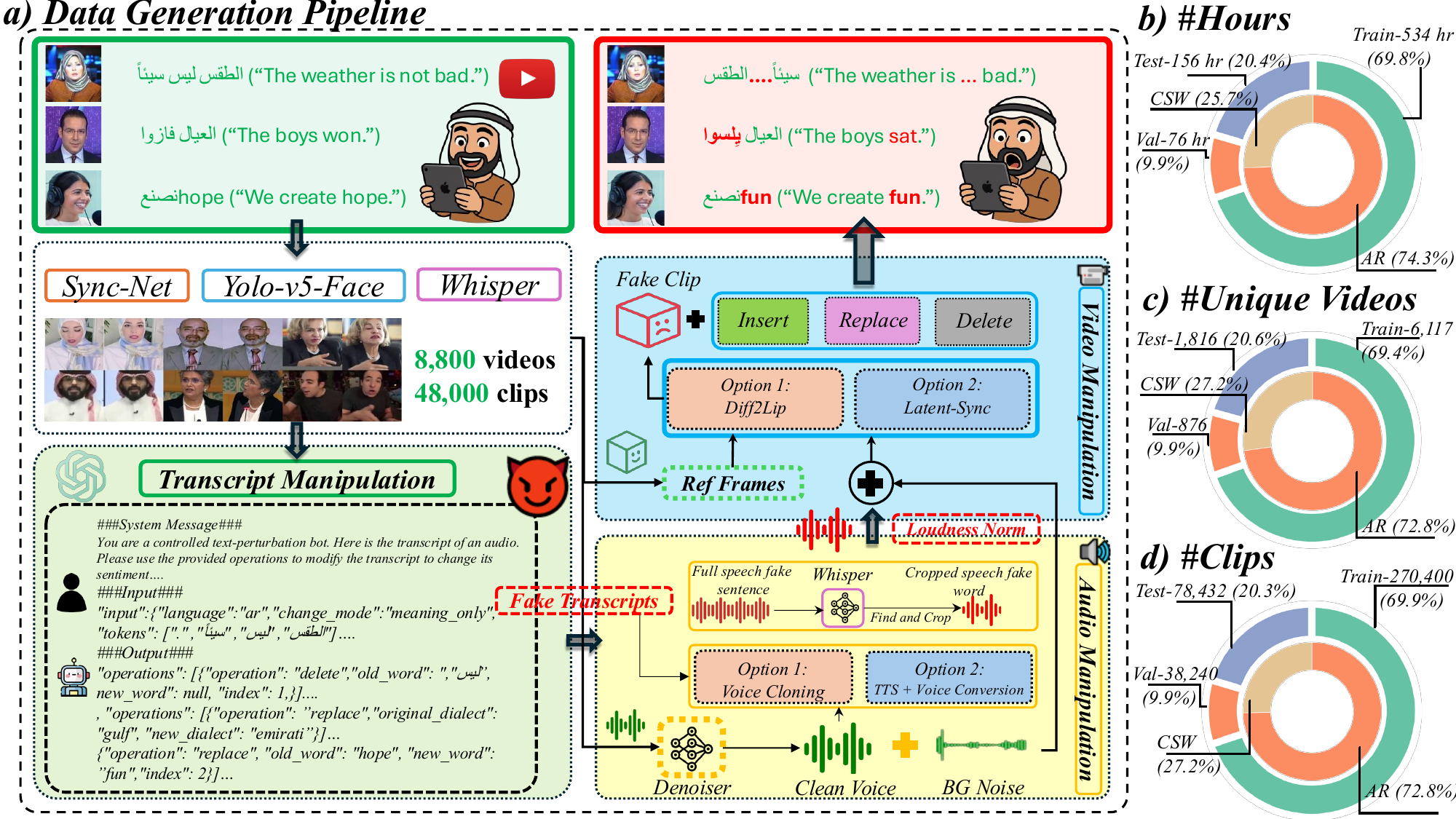}
  \caption{ a) We show the data generation pipeline for ArEnAV dataset. In a) input videos are analysed for audio, face, and text extraction. Using few-shot prompts with GPT-4.1-mini, CSW-based spoken text manipulation is performed. This is followed by speech and face enactment generation. b-d) The plots show the data splits and CSW distribution. Here is an example of CSW 
  input and 
  manipulated text
  with translations in parentheses: 
  \<نصنع> hope (“We create hope.”) --> \<نصنع> fun (“We create fun.”)}
  \label{fig:teaser}
\end{figure*}

\begin{figure*}[b]
  \centering
  \includegraphics[width=\textwidth]{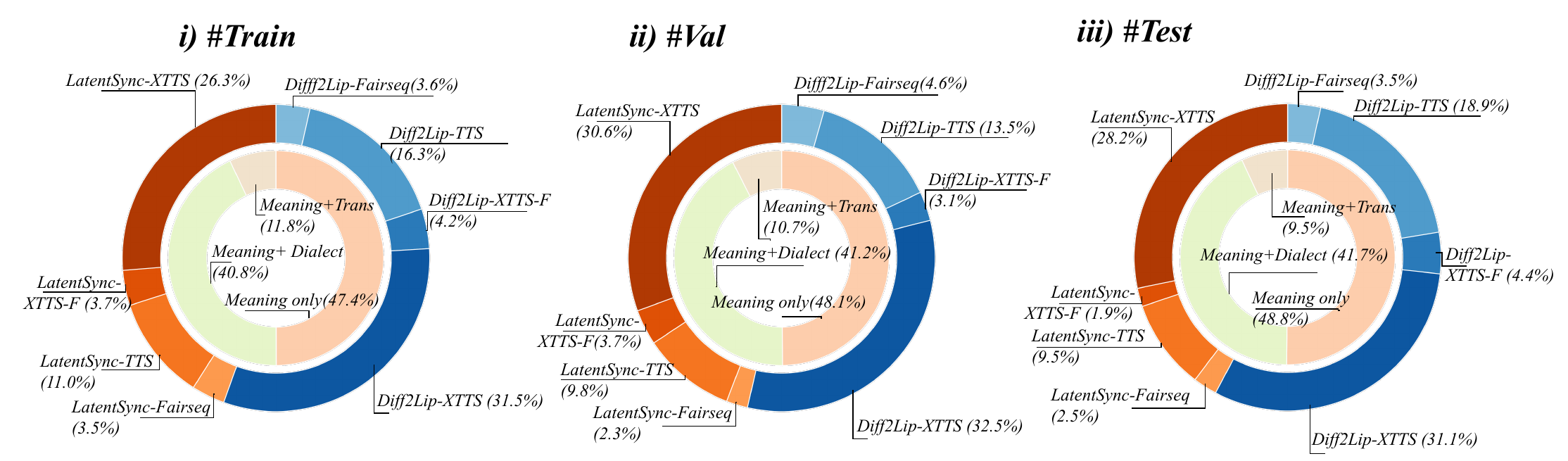}
  \caption{Dataset distribution for i) \emph{Train}, ii) \emph{Val} and iii) \emph{Test} split. The outer layer shows the split between various combinations of Text-to-Speech and Lip-Sync models used for audio-visual manipulation. The middle layer shows the distribution of language in the original transcript, which is either \emph{Ar} (Arabic) or \emph{CSW} (Code-Switched English-Arabic).  The inner layer shows the distribution of different operations applied to the original transcripts, "meaning only", "dialect+meaning", and "meaning + translation" (For fine-grained detail about what they entail, refer to Table \ref{tab:augmentation_examples}.)}
  \label{fig:transcripts_eval}
\end{figure*}

\begin{table*}[t]
  \centering
  \captionsetup[table]{skip=3pt}
  \caption{\textbf{Details for publicly available deep‑fake datasets in chronologically ascending order.}
           \textmd{Cla: Binary classification, SL: Spatial localization, TL: Temporal localization,
           FS: Face swapping, RE: Face reenactment, TTS: Text‑to‑speech, VC: Voice conversion.}}
  \label{tab:existing_datasets}

  \setlength{\tabcolsep}{5pt}

  \resizebox{\textwidth}{!}{%
  \begin{tabular}{l||c|c|c|c|c|c|c}
    \toprule[0.4mm]
    \rowcolor{mygray}\textbf{Dataset} & \textbf{Year} & \textbf{Tasks} &
      \textbf{Manip.\ Modality} & \textbf{Method} &
      \textbf{\#Total} & \textbf{Multilingual} & \textbf{Code Switching} \\
    \midrule\midrule
    Google DFD~\citet{nickContributing2019} & 2019 & Cla &
      V & FS & 3,431 & \xmark& \xmark\\
    DFDC~\citet{dolhanskyDeepFake2020} & 2020 & Cla &
      AV & FS & 128,154 & \xmark& \xmark\\
    DeeperForensics~\citet{jiangDeeperForensics12020} & 2020 & Cla &
      V & FS & 60,000 & \xmark& \xmark\\
    Celeb\textendash DF~\citet{liCelebDF2020} & 2020 & Cla &
      V & FS & 6,229 & \xmark& \xmark\\
    KoDF~\citet{kwonKoDF2021} & 2021 & Cla &
      V & FS/RE & 237,942 & \xmark& \xmark\\
    FakeAVCeleb~\citet{khalid2022fakeavcelebnovelaudiovideomultimodal} & 2021 & Cla &
      AV & RE & 25,500$+$ & \xmark& \xmark\\
    ForgeryNet~\citet{heForgeryNet2021} & 2021 & SL/TL/Cla &
      V & Random FS/RE & 221,247 & \xmark& \xmark\\
    ASVSpoof2021DF~\citet{liuASVspoof2023} & 2021 & Cla &
      A & TTS/VC & 593,253 & \xmark& \xmark\\
    LAV\textendash DF~\citet{caiYou2022} & 2022 & TL/Cla &
      AV & Content‑driven RE/TTS & 136,304 & \xmark& \xmark\\
    DF\textendash Platter~\citet{narayanDFPlatter2023} & 2023 & Cla &
      V & FS & 265\,756 & \xmark& \xmark\\
    AV-1M\citet{caiAVDeepfake1M2023} & 2023 & TL/Cla &
      AV & Content‑driven RE/TTS & 1,146,760 & \xmark& \xmark\\[1pt]
PolyGlotFake\citet{hou2024polyglotfakenovelmultilingualmultimodal}   & 2024  & Cla  & AV & RE/TTS/VC   & 15,238 & \cmark  & \xmark\\[-1pt]
Illusion\citet{thakral2025illusion}   & 2025  & Cla  & AV & FS/RE/TTS   & 1,376,371 & \cmark  & \xmark\\[-1pt]

     \textbf{ArEnAV (Ours)}& \textbf{2025}  & \textbf{Cla/TL}  & \textbf{AV} & \textbf{Content Driven RE/TTS/VC}  & \textbf{387,072}  & \cmark  & \cmark   \\[2pt]
    \bottomrule[0.4mm]
  \end{tabular}}
\end{table*}

\vspace{-1mm}
\section{Related Work}

Early deepfake research was predominantly monolingual and modality-specific. Initial significant contributions included video manipulation techniques such as FaceSwap and Face2Face as introduced by \citet{thies2020face2facerealtimefacecapture}, which led to seminal datasets like FaceForensics++ (\citet{rössler2019faceforensicslearningdetectmanipulated}) and the DeepFake Detection Challenge (DFDC) (\citet{dolhansky2020deepfakedetectionchallengedfdc}). These datasets primarily provided facial manipulations within single-language contexts, focusing largely on visual realism.

Parallel to video deepfake advancements, audio deepfakes evolved rapidly, driven by progress in text-to-speech (TTS) synthesis, voice conversion, and generative audio models such as Tacotron (\citet{wang2017tacotronendtoendspeechsynthesis}). Datasets like ASVspoof (\citet{wang2020asvspoof2019largescalepublic}) and WaveFake (\citet{frank2021wavefakedatasetfacilitate}) contributed significantly by providing benchmarks to evaluate audio manipulation detection methods, albeit still largely restricted to English.

In recent years, research has expanded towards joint audio-visual deepfake manipulations. Datasets such as FakeAVCeleb (\citet{khalid2022fakeavcelebnovelaudiovideomultimodal}) showcased realistic lip-synced speech synthesis in tandem with facial manipulations. AV-Deepfake1M (\citet{cai20241mdeepfakesdetectionchallenge}) further advanced this domain by automating transcript alterations to create nuanced, localized audio-visual deepfakes, highlighting the necessity of detecting temporally and contextually subtle manipulations.

Recently, there has been increased focus on multilingual audio deepfakes. These efforts have revealed key limitations in generalizing detection models across languages and proposed new resources to address these challenges. \citet{marek2024audiodeepfakedetectionmodels} conducted a comprehensive study on cross-lingual deepfake detection, showing that models trained on one language often fail to generalize effectively to others, underscoring the role of language-specific phonetic and prosodic features in model performance. 
Complementing these dataset-focused efforts, \citet{phukan2024heterogeneityhomogeneityinvestigatingmultilingual} explored the utility of multilingual pre-trained models (PTMs) for detection tasks, finding that such models capture language-agnostic representations that improve cross-lingual robustness. Multilingual audio-visual datasets emerged even more recently to address the global dimension of deepfake threats. The PolyGlotFake dataset (\citet{hou2024polyglotfakenovelmultilingualmultimodal}) contains audio-visual deepfakes across seven languages. Although the dataset covers a wide range of language, the size of the real data is significantly small.  
Nonetheless, these multilingual datasets remain limited to either monolingual or monomodal scenarios within each single instance, ignoring the prevalent reality of intra-utterance language switching.

Our work directly responds to this critical gap. Unlike previous studies, we not only incorporate multilingual content but also explicitly generate intra-utterance code-switched audio-visual deepfakes. We leverage SOTA TTS and lip sync methodologies adapted for multilingual use, resulting in realistic, diverse, and challenging benchmarks.

\section{ArEnAV Dataset} ArEnAV is a large-scale audio-visual deepfake dataset specifically focused on Arabic–English CSW. 
Comprising approximately 765 hours of video data sourced from 8,809 unique YouTube videos, ArEnAV establishes itself as the first and most extensive benchmark for multilingual deepfake detection (see Table \ref{tab:existing_datasets} for dataset comparison). 
The dataset is constructed to preserve the original identity and environmental context of the source videos while systematically manipulating the semantic content to introduce Arabic-English CSW. 
Following the taxonomy proposed by \citet{cai20241mdeepfakesdetectionchallenge}, ArEnAV includes three 
manipulation strategies:
 \textbf{Fake Audio \& Fake Video}: Both audio and visual content are synthetically generated, simulating complete audiovisual deepfakes.
 \textbf{Fake Audio \& Real Video}: The audio track is manipulated to introduce anti-semantic and CSW content while maintaining the original visual content.
 \textbf{Real Audio \& Fake Video}: The original audio is retained, while facial movements and lip synchronization are altered to create visually deceptive content.

\subsection{Data Collection}
We use the YouTube video links from VisPer's Arabic Train subset (\citet{narayan2024vispermultilingualaudiovisualspeech}). Further, we first run a scene change detection model to split the video into clips, and then we use Yolo-v5 to obtain the faces in each frame as well as track them across frames. Afterwards, we generate the ground truth transcripts using a Whisper-V2 model introduced by \citet{radford2022robustspeechrecognitionlargescale}, further fine-tuned on Arabic-English code-switching data, with the default output language set to Arabic. Following the transcripts, we apply Forced Alignment between the audio and text, using a multilingual wav2vec2 model (\citet{baevski2020wav2vec20frameworkselfsupervised}) supporting both Arabic and English. This provides us with word-level timestamps for code-switched Arabic and English data. \vspace{-3mm}

\subsection{Data Generation Pipeline} The data generation pipeline roughly consists of three stages: transcript manipulation, audio generation, and video generation. First, we obtain an accurate transcription of each source video and apply controlled modifications to the text. Next, we synthesize new audio for the altered transcript while preserving the speaker’s voice characteristics. Finally, we render a lip-synced video that matches the new audio, producing a realistically manipulated video clip. We detail each stage as follows.

\begin{table}[b] \vspace{-2mm}
  \centering
  \caption{Transcript manipulation rules in ArEnAV for Arabic (AR) and English (EN) words.}
  \label{tab:manipulation_rules}
  \resizebox{\linewidth}{!}{%
  \begin{tabular}{@{}c c c c p{8cm}@{}}
    \toprule
    \rowcolor{mygray}
    \# & \textbf{Original Transcript} & \textbf{Original Word} & \textbf{Inserted Word} & \textbf{Operation} \\ 
    \midrule \midrule
    1 & CSW & EN & EN & Change meaning only (keep English) \\
    2 & CSW & AR & AR & Change meaning only (keep Arabic variant) \\
    3 & CSW & AR & AR & Change meaning + change Arabic variant \\
    4 & CSW & AR & AR/EN & multi-op; When 2-3 ops → edit 1 EN and 1-2 AR words \\
    \midrule
    5 & Arabic & AR & AR & Change meaning only (keep Arabic variant) \\
    6 & Arabic & AR & AR & Change meaning + change Arabic variant \\
    7 & Arabic & AR & EN & Change meaning + change language to English \\
    8 & Arabic & AR & AR/EN & multi-op; Apply all operations on Arabic words\\
    \bottomrule
  \end{tabular}
  }
  \vspace{-3mm}
\end{table}

\subsubsection{Transcript Manipulation}
\label{sec:transcript_manipulation}
We leverage GPT-4.1-mini (\citet{openai_gpt41_mini}) to perform content-driven modifications of our multilingual transcripts. We define eight distinct transcript change modes that span both code-switched and Arabic-only contexts, allowing fine-grained control over how the transcript is altered. These modes include three main operations: first, \textit{ meaning only}, which only involves changing the meaning of the word and keeping the language as it is, second, \textit{ meaning + dialect }, which involves changing the meaning of the word and changing its language to 
another Arabic variant (either MSA or any dialect), and lastly, \textit{meaning + translation}, which asks the model to change the meaning of the word, and then translate it to English. Table~\ref{tab:manipulation_rules} summarizes the eight modification modes with their intended effect. By categorizing edits in this way, we ensure a controlled and diverse set of manipulations ranging from subtle word substitutions to introducing or removing CSW instances. Due to the effectiveness of few-shot prompting, we prompt GPT-4.1-mini with 15 examples, explaining various kinds of transitions and  possible changes. 
Examples of original and augmented transcripts achieved by these manipulation rules are shown in Appendix~\ref{sec:appendix_aug_examples}. We provide the prompt in Appendix~\ref{sec:prompt}. 
We report text manipulations distributions as follows: replacement (94.6\%), insertion (5.1\%) and deletion (0.3\%). Please note that this distribution was randomly chosen based on input text statements by GPT.

 \textbf{
\textit{\subsubsubsection{Transcript Quality:}}}

To quantify the impact of our LLM-based manipulations, we employ two complementary metrics:
\textbf{\textit{Bidirectional Entailment Quality Mean}}: the average of Real→Fake and Fake→Real NLI entailment scores (1.0 = full semantic entailment; 0.0 = direct contradiction) and 
\textbf{\textit{Perplexity}}: how well a language model predicts a transcript (lower = more fluent/natural).
Table \ref{tab:entailment} shows the distribution of entailment quality means over different types of perturbations (meaning only, dialect + meaning, meaning + translation). In every subset, a large fraction of samples lies below the 0.5 threshold, and many even in the contradiction zone, demonstrating that our pipeline reliably injects semantic change regardless of language or dialect.

\begin{table}[t]
  \centering
  \caption{a) Entailment distribution over i) All change modes, ii) Dialect + meaning, iii) Meaning only, and iv) Meaning + translation. b) Perplexity Evaluation distribution among dataset splits, showing perplexity calculated using i) Jais-3B, an Arabic-English LLM, and ii) Qwen-2.5-7B. 
  Reference in both shows the perplexity calculated on an Arabic-English CSW text dataset(\citet{sdaian_cai_code_switching_dataset}). }
  \begin{subtable}[t]{0.48\columnwidth}
    \centering
    \small
    \caption{Entailment distribution.}
    \resizebox{\linewidth}{!}{%
      \begin{tabular}{l|ccccc}
        \toprule
        \rowcolor{gray!25}
        \textbf{Subset} & \textbf{0–0.2} & \textbf{0.2–0.4} & \textbf{0.4–0.6} & \textbf{0.6–0.8} & \textbf{0.8–1.0} \\
        \midrule
        Total                  & 32011 & 4160 & 3952 & 3476 & 4785 \\
        Dialect + Meaning      & 12133 & 2154 & 2126 & 1891 & 2473 \\
        Meaning Only           & 19107 & 1765 & 1463 &  967 &  882 \\
        Meaning + Translation  &   771 &  241 &  363 &  618 & 1430 \\
        \bottomrule
      \end{tabular}%
    }
    \label{tab:entailment}
  \end{subtable}\hfill%
  \begin{subtable}[t]{0.48\columnwidth}
    \centering
    \small
    \caption{Perplexity Evaluation distribution.}
    \resizebox{\linewidth}{!}{%
      \begin{tabular}{l|cc|cc|cc|c}
        \toprule
        \rowcolor{gray!25}
        \textbf{Model}   & \multicolumn{2}{c|}{\textbf{Train}}   & \multicolumn{2}{c|}{\textbf{Val}}     & \multicolumn{2}{c}{\textbf{Test}}  & \textbf{Reference}   \\
        \midrule
        \rowcolor{gray!25}
                & \textbf{Real}      & \textbf{Fake}         & \textbf{Real}      & \textbf{Fake}         & \textbf{Real}      & \textbf{Fake}  &       \\
        \midrule
        Jais-3B & 1207.26   & 1211.39      &  997.39   & 1001.45      & 1146.39   & 1144.14  &324.1    \\
        Qwen-7B &  116.58   &  117.44      &  124.66   &  125.07      &  117.45   &  117.58  & 77.44    \\
        \bottomrule
      \end{tabular}%
    }
   
  \end{subtable}
   \label{tab:Perpexity}
\end{table}

Table \ref{tab:Perpexity} reports average perplexities on real versus fake transcripts under two open-source LLMs; Jais-3B (\citet{sengupta2023jaisjaischatarabiccentricfoundation}) and Qwen-2.5-7B (\citet{qwen2025qwen25technicalreport}), across the data splits. The minimal difference in perplexity shows that our fake transcripts remain fluent and natural, despite major changes in meaning. This balance between altered content and surface-level fluency is essential for generating effective audio-visual deepfakes

\subsubsection{Audio Generation}
The next step involves generating a synthetic audio track that precisely follows the edited transcript while maintaining the voice characteristics of the original on-screen speaker. Initially, we segment the audio into clean speech and background noise using a Denoiser (\citet{defossez2020real}). Conventional zero-shot voice cloning systems, such as YourTTS (\citet{casanova2023yourttszeroshotmultispeakertts}) exhibit strong performance in English but struggle with Arabic phonetics and cross-lingual synthesis. To address this, we employ four targeted cloning strategies:
a) \textbf{XTTS-v2 (\citet{casanova2024xttsmassivelymultilingualzeroshot})}: A multilingual, zero-shot TTS model natively supporting Arabic, English, and code-switching.
b) \textbf{XTTS-v2 (\citet{casanova2024xttsmassivelymultilingualzeroshot}) + OpenVoice-v2 (\citet{qin2024openvoiceversatileinstantvoice})}: When a reference voice sample is available, we achieve higher fidelity by generating the utterance with XTTS-v2 and performing speaker conversion via OpenVoice-v2.
c) \textbf{Fairseq Arabic TTS (\citet{ott-etal-2019-fairseq}) + OpenVoice-v2 (\citet{qin2024openvoiceversatileinstantvoice})}: For fully Arabic sentences, we generate audio with the Fairseq Arabic TTS from the MIMS initiative, followed by speaker conversion using OpenVoice-v2.
d) \textbf{GPT-TTS (\citet{openai_tts1}) + OpenVoice-v2 (\citet{qin2024openvoiceversatileinstantvoice})}: GPT-TTS supports 29 languages with nine preset voices; we randomly select one voice, generate the sentence, and then convert the audio to the target speaker's voice with OpenVoice-v2.

The audio-generation flow depends on the edit type. For \emph{insert} or \emph{replace} operations, we regenerate the complete sentence and validate the generated audio using Whisper-Turbo (\citet{radford2022whisper}), retaining only samples that exactly match the intended transcript. This step ensures intelligibility and accurate timestamp alignment for splicing the segment into the original audio. If validation fails, we discard the sample. For a \emph{delete} operation, we remove speech segments entirely, preserving only background noise. Finally, after each edit, we normalise the loudness of the manipulated segment relative to the original audio and recombine it with the extracted environment noise. 

\subsubsubsection{\textit{\textbf{Audio Quality:}}} 

\begin{wraptable}{r}{0.45\columnwidth}
 \caption{Audio quality comparison across different datasets.}
  \vspace{-3mm}
  \scriptsize
  \setlength\tabcolsep{2pt}
  \begin{tabular}{@{}l c c c c@{}}
    \toprule
    \rowcolor{gray!25}
    Dataset         & Language       & SECS$\uparrow$ & SNR(dB)$\uparrow$ & FAD$\downarrow$ \\
    \midrule
    FakeAVCeleb     & English        & 0.543           & 2.16                & 6.598            \\
    LAV-DF          & English        & 0.984           & 7.83                & 0.306            \\
    AV-Deepfake1M   & English        & 0.991           & 9.39                & 0.088            \\
    \midrule
    \textbf{ArEnAV} & Arabic, English & \textbf{0.990}  & 7.65 & \textbf{0.140}   \\
    \bottomrule
  \end{tabular}
 
  \label{tab:audio_quality}
  \vspace{-2mm}
\end{wraptable}
Table~\ref{tab:audio_quality} presents the comparison of audio quality for ArEnAV based on speaker similarity, signal quality and distribution realism with existing audio-visual deepfake datasets. We report speaker encoder cosine similarity (SECS), Signal-to-Noise (SNR) and Fréchet audio distance (FAD) for recent Audio-Visual datasets. SECS measures the speaker’s voice similarity between a generated clip and the real reference (range $[-1,1]$, higher is better), while FAD evaluates the distributional distance between the generated audio and real audio (lower is better). The metrics combined indicate that ArEnAV has high-quality audio samples.

\begin{table}[ht] \vspace{-1.5mm}
  \centering
  \scriptsize
  \setlength\tabcolsep{2pt}

  \caption{(a) Visual quality comparison across different datasets and (b) Distribution of human‐reported reasons for labellng a video as fake in the user study.}
  \begin{subtable}[t]{0.48\columnwidth}
    \centering
    \footnotesize
    \caption{Visual quality metrics.}
    \label{tab:video_quality_sub}
    \resizebox{\linewidth}{!}{%
      \begin{tabular}{@{}l c c c@{}}
        \toprule
        \rowcolor{gray!25}
        Dataset         & PSNR(dB)$\uparrow$ & SSIM$\uparrow$ & FID$\downarrow$ \\
        \midrule
        FF++            & 24.40       & 0.812  & 1.06  \\
        DFDC            & —           & —      & 5.69  \\
        FakeAVCeleb     & 29.82       & 0.919  & 2.29  \\
        LAV-DF          & 33.06       & 0.898  & 1.92  \\
        AV-Deepfake1M   & 39.49       & 0.977  & 0.49  \\
        \midrule
        \textbf{ArEnAV} & \textbf{37.70} & \textbf{0.971} & \textbf{0.68} \\
        \bottomrule
      \end{tabular}%
    }
  \end{subtable}\hfill
  \begin{subtable}[t]{0.48\columnwidth}
    \centering
    \caption{Distribution of top reasons.}
    \label{tab:reason_distribution_sub}
    \resizebox{\linewidth}{!}{%
      \begin{tabular}{@{}l r@{}}
        \toprule
        \rowcolor{gray!25}
        Reason                            & Percentage (\%) \\
        \midrule
        Unintelligible speech (weird audio)    & 36.5 \\
        Video/audio mismatch (lip sync is off) & 25.1 \\
        Audio sounds artificial                & 24.7 \\
        Video looks artificial                 &  8.7 \\
        Code-switching is unnatural            &  3.0 \\
        Incoherent sentence                    &  1.9 \\
        \bottomrule
      \end{tabular}%
    }
  \end{subtable}

  \label{tab:visual_and_reason}
\end{table}

\subsection{Visual Manipulation} 

For video generation, after extensive experimentation, we chose two diffusion-based lip-sync approaches: Diff2Lip (\citet{mukhopadhyay2023diff2lipaudioconditioneddiffusion}) and LatentSync (\citet{li2025latentsynctamingaudioconditionedlatent}). Both of these models perform high quality zero-shot lip-sync and are open-sourced. Using the new generated audio and original video's frames we generate the fake frames. For \textit{replace} and \textit{insert} word operations, we generate the fake frames for the new word, and for \textit{delete} word operations, we generate a face with closed lips i.e. without audio.

\begin{table}[b]
\centering
\small                          
\caption{Data distribution in ArEnAV and comparison with other multilingual datasets.}
\resizebox{0.9\columnwidth}{!}{               
  \begin{tabular}{l|ccccccc}
    \toprule[0.2mm]
    \rowcolor{mygray}
    \textbf{Subset} & 
    \begin{tabular}{@{}c@{}}\textbf{\#Unique} \\ \textbf{Videos}\end{tabular} & 
    \begin{tabular}{@{}c@{}}\textbf{\#Real} \\ \textbf{Videos}\end{tabular}& 
    \begin{tabular}{@{}c@{}}\textbf{\#Fake} \\ \textbf{Videos}\end{tabular} &
    \begin{tabular}{@{}c@{}}\textbf{\#Non-English} \\ \textbf{Clips}\end{tabular} &
    \begin{tabular}{@{}c@{}}\textbf{\#CSW} \\ \textbf{Videos}\end{tabular} &
    \begin{tabular}{@{}c@{}}\textbf{\#Arabic} \\ \textbf{Videos}\end{tabular} &
    \begin{tabular}{@{}c@{}}\textbf{Arabic} \\ \textbf{Variants}\end{tabular}\\
    \midrule
    PolyGlotFake \cite{hou2024polyglotfakenovelmultilingualmultimodal} 
      & 766   & 766     & 14,472   & \textbf{11,941} & \textbf{0}   & \textbf{1,403} & NA \\
    Illusion \cite{thakral2025illusion} 
      & –     & 141,440 & 1,234,931& \textbf{4,385}  & \textbf{0}   & –   &      NA    \\
    \midrule
    \textbf{ArEnAV-Train}      
      & 6,117 & 67,600  & 202,800  & 270,400         & 69,544       & 200,856     &  Egyptian,  \\
    \textbf{ArEnAV-Validation} 
      & 876   & 9,560   & 28,680   & 38,240          & 10,416       & 27,824       &  MSA,  \\
    \textbf{ArEnAV-Test}       
      & 1,816 & 19,608  & 58,824   & 78,432          & 19,832       & 58,600      &  Levantine, Gulf   \\ 
    \midrule
    \textbf{ArEnAV (total)}    
      & \textbf{8,809} & \textbf{96,768}  & \textbf{290,304}  & \textbf{387,072}         & \textbf{99,792}       & \textbf{287,280 }    &  - \\
    \bottomrule[0.1mm]
  \end{tabular}
}

\label{tab:dataset_stats}
\end{table}

\textbf{\textit{Visual Quality:}}

To evaluate visual quality, we use three standard metrics: Peak Signal-to-Noise Ratio (PSNR), Structural Similarity Index (SSIM), and Fréchet Inception Distance (FID). Table~\ref{tab:visual_and_reason} presents PSNR, SSIM, and FID results for the ArEnAV dataset.
PSNR and SSIM measure pixel-level and structural similarity, respectively, between fake and original frames (higher is better) (ArEnAV lies nearby AV-1M). FID assesses realism by comparing the distributions of fake and real frames in a learned image feature space (lower is better) (ArEnAV slightly more than AV-1M). These scores highlight that ArEnAV attains higher / comparable visual quality compared to other deepfake datasets. 
\textbf{Real Perturbations:}
To mimic real-life videos scenarios better, we add localized perturbations to both the real and the fake videos. We apply 15 different visual filters (eg: salt-pepper noise and camera shaking) and 10 different audio manipulations (eg: time-stretching, random loudness and pitch). For each video, we randomly sample one to three instances for visual perturbations and one to two instances for audio perturbations. Perturbation details are mentioned in Appendix \ref{lab:Appendix}.

\subsection{User Study}\label{section: user_study}

\begin{wraptable}{r}{0.5\columnwidth}
  \caption{User study results show that the deepfake detection and localization in multilingual CSW videos is non-trivial for human observers.}
    \centering
  \vspace{-3mm}
  \scriptsize
  \setlength\tabcolsep{2pt}
  \begin{tabular}{@{}l|cccc@{}}
    \toprule
    \rowcolor{gray!25}
    Method & Acc.\ & AP@0.1 & AP@0.5 & AR@1 \\
    \midrule
    ArEnAV & 60.00  & 8.35    & 0.79    & 1.38  \\
    \bottomrule
  \end{tabular}
  \label{tab:user_study}
  \vspace{-1.5mm}
\end{wraptable}

To investigate whether humans can identify deepfakes in ArEnAV, we conducted a user study with 19 participants, out of which, 15 are native Arabic speakers, and 4 have basic knowledge of Arabic. We randomly sampled 20 videos, with either 0 or 1 manipulation. \textbf{\textit{Instructions for User Study:}} Each participant was asked to 1) watch the video, and 2) answer 3 questions, i) Is the video real of fake, ii) If it is fake, localize where they think the fake region is, iii) Whether the given video contains Arabic-English code-switching or not, iv) Give a reason for labelling the video (if they have) as a deepfake. The results in Table \ref{tab:user_study} reaffirm our hypothesis that identifying audiovisual deepfakes in multilingual (specially CSW) and multimodal settings is a non-trivial task, as even humans achieve only 60\% accuracy, while it is even harder to localize the deepfakes, with AP@0.5 at 0.79. Further, Table \ref{tab:reason_distribution_sub} shows the primary reasons why people classified the videos as fake. We report that 85\% of the users fail to identify deepfakes when the manipulation happens in the English word, in the CSW video, which can be attributed to a higher quality of voice cloning in English as well as the natural change in tone when a person code-switches, which makes it harder to detect. Further, localization is very tough due to the very high quality of lip-sync with diffusion models, as shown in Table \ref{tab:reason_distribution_sub}, where the video being the reason for fake classification is only 8.7\%.

\subsection{Dataset Statistics} \label{ComputeandStats}
Table \ref{tab:dataset_stats} compares ArEnAV with other multilingual deepfake detection datasets. Existing multilingual datasets like PolyGlotFake (\citet{hou2024polyglotfakenovelmultilingualmultimodal}) and Illusion (\citet{thakral2025illusion}) have significantly smaller multilingual content, containing limited Arabic data (1,400 Arabic videos in PolyGlotFake and minimal in Illusion across 26 languages). ArEnAV includes 387k videos sourced from 8,809 unique YouTube videos, totaling over 765 hours. Videos average approximately 7.7 seconds each, with train, validation, and test splits created via multilabel stratified sampling in a 7:1:2 ratio, ensuring no overlap.

\textbf{ Computational Cost:} \label{subsection: comp_cost} We spent around ~50 GPU hours to generate the real transcript using
Whisper-Large-V2 (\citet{radford2022whisper}), 200 dollars worth of OpenAI credits, to generate fake transcripts and Text-to-Speech model, TTS-1 (\citet{openai_tts1}), and 650 GPU hours for
data generation. Overall, we needed ~800 GPU hours to generate
AvEnAV with NVIDIA RTX-\emph{6000} GPUs.

\section{Benchmark and Metrics}

\newcolumntype{Y}{>{\centering\arraybackslash}X}

\begin{table}[b] \vspace{-1.5em}
  \caption{Temporal localization results on the test set of ArEnAV.}
  \centering
  {\footnotesize
  \setlength{\tabcolsep}{2pt}
  \scalebox{0.9}{
  \begin{tabularx}{\linewidth}{c|l|c|*{4}{>{\centering\arraybackslash}X}|
                                       *{5}{>{\centering\arraybackslash}X}}
    \toprule[0.4mm]
    \rowcolor{mygray}
    \textbf{Set} & \textbf{Method} & \textbf{Mod.} &
    \textbf{AP@0.5} & \textbf{AP@0.75} & \textbf{AP@0.9} & \textbf{AP@0.95} &
    \textbf{AR@50} & \textbf{AR@30} & \textbf{AR@20} & \textbf{AR@10} & \textbf{AR@5} \\[2pt]
    \midrule
    \multirow{6}{*}{\rotatebox[origin=c]{90}{\textbf{Full dataset}}}
      & Meso4  & V  & 0.02 & 0.01 & 0.00 & 0.00 & 0.09 & 0.09 & 0.09 & 0.09 & 0.09 \\
      & MesoInception  & V & 0.56 & 0.18 & 0.04 & 0.01 & 4.11 & 4.11 & 4.11 & 4.11 & 4.08 \\
      & Xception & V  & 22.50 & 10.26 & 2.29 & 0.58 & 19.13 & 19.13 & 19.13 & 19.13 & 19.13 \\
    & BA-TFD (ZS) & AV  &0.17 & 0.01 & 0.00 & 0.00 & 9.72 & 5.20 & 3.07 & 1.46 & 0.73\\
      & BA-TFD+ (ZS) & AV &0.11 & 0.00 & 0.00 & 0.00 & 5.77 & 2.95 & 2.09 & 0.87 & 0.37  \\
      & BA-TFD & AV  &2.42 & 0.55 & 0.01 & 0.00 & 22.30 & 10.31 & 3.41 & 2.54 & 1.67  \\
      & BA-TFD+ & AV  &3.74 & 1.10 & 0.06 & 0.01 & 30.75 & 9.42 & 4.55 & 3.05 & 1.83 \\

    \midrule
    \multirow{6}{*}{\rotatebox[origin=c]{90}{\textbf{Set V}}}
      & \multicolumn{11}{l}{} \\[-6pt]  
      & Meso4  & V  & 0.02 & 0.01 & 0.00 & 0.00 & 0.10 & 0.10 & 0.10 & 0.10 & 0.10 \\
      & MesoInception  & V  & 0.83 & 0.27 & 0.05 & 0.01 & 5.56 & 5.56 & 5.56 & 5.56 & 5.53 \\
      & Xception & V & 32.76 & 14.48 & 3.30 & 0.81 & 27.78 & 27.78 & 27.78 & 27.78 & 27.78 \\
    & BA-TFD (ZS) & AV  &0.12 & 0.00 & 0.00 & 0.00 & 8.44 & 4.34 & 2.44 & 1.13 & 0.49  \\
      & BA-TFD+ (ZS) & AV &0.07 & 0.00 & 0.00 & 0.00 & 4.69 & 2.39 & 1.65 & 0.69 & 0.29 \\
      & BA-TFD & AV  &3.65 & 0.25 & 0.01 & 0.00 & 25.31 & 9.03 & 3.64 & 2.34 & 1.64  \\
      & BA-TFD+ & AV  &5.65 & 1.89 & 0.08 & 0.02 & 31.09 & 13.21 & 5.91 & 3.05 & 2.05 \\
    \midrule
    \multirow{6}{*}{\rotatebox[origin=c]{90}{\textbf{Set A}}}
      & \multicolumn{11}{l}{} \\[-6pt]
      & Meso4  & V  & 0.02 & 0.01 & 0.00 & 0.00 & 0.08 & 0.08 & 0.08 & 0.08 & 0.08 \\
      & MesoInception  & V  & 0.38 & 0.09 & 0.01 & 0.00 & 3.25 & 3.25 & 3.25 & 3.25 & 3.22 \\
      & Xception & V  & 14.72 & 3.92 & 0.29 & 0.09 & 11.78 & 11.78 & 11.78 & 11.78 & 11.78 \\
    & BA-TFD (ZS) & AV  &0.23 & 0.01 & 0.00 & 0.00 & 12.14 & 6.46 & 3.85 & 1.83 & 0.95 \\
      & BA-TFD+ (ZS) & AV &0.14 & 0.01 & 0.00 & 0.00 & 7.32 & 3.79 & 2.69 & 1.13 & 0.48  \\
      & BA-TFD & AV  &3.21 & 0.60 & 0.02 & 0.00 & 24.45 & 9.26 & 4.15 & 2.61 & 1.93  \\
      & BA-TFD+ & AV  &4.35 & 1.10 & 0.10 & 0.00 & 28.35 & 11.23 & 4.85 & 3.11 & 2.00 \\
    \bottomrule[0.2mm]
  \end{tabularx}}}

  \label{tab:temporal_localization}
\end{table}

We organize the data into \emph{train}, \emph{validation}, and \emph{test} split. We use multilabel stratified sampling to divide the data in equal proportions based on the method type, the change mode, and the ground truth language. We also show evaluation on two subsets, \emph{subset V}, which excludes videos with audio-only manipulation, and \emph{subset A}, which excludes videos with visual-only manipulations. We evaluate models on two tasks, \textbf{temporal localization} and \textbf{detection} of audio-visual deepfakes. We use average precision (AP) and average recall (AR) metrics as prior works~(\citet{heForgeryNet2021, caiYou2022,caiAVDeepfake1M2023}) for temporal localization. For the task of deepfake detection, we use standard evaluation protocol (\citet{rosslerFaceForensics2019, dolhanskyDeepFake2020,caiAVDeepfake1M2023}) to report video-level accuracy (Acc.) and area under the curve (AUC).

\noindent\textbf{Implementation Details:}\label{sec: implementation_details}
We benchmark \textbf{temporal detection} using SOTA models: Meso4, MesoInception4, Xception, BA-TFD, and BA-TFD+. BA-TFD and BA-TFD+ (\citet{caiGlitch2023}) are evaluated in their original configurations, both in a zero-shot setting (pre-trained on AV-1M; \citet{caiAVDeepfake1M2023}) and after fine-tuning on our dataset. For image-based classifiers (Meso4, MesoInception4; \citet{afcharMesoNet2018} and Xception; \citet{cholletXception2017}), we aggregate frame-level predictions to segments following \citet{caiAVDeepfake1M2023}. 
For benchmarking \textbf{deepfake detection}, image-based models (Meso4, MesoInception4, and Xception) are trained on video frames with corresponding labels, and predictions are aggregated to video-level using max voting, as suggested by \citet{caiAVDeepfake1M2023}. Additionally, we assess zero-shot performance of LLM-based models, VideoLLaMA2 and VideoLLaMA2.1-AV (\citet{zhangVideoLLaMA2023}), prompting them to produce a confidence score indicating the likelihood of a video being a deepfake. We include an audio-only baseline, XLSR-Mamba (\citet{xiao2025xlsrmambadualcolumnbidirectionalstate}), the best open-source audio deepfake detection model on Speech DF Arena (\citet{huggingface_speech_df_arena}), evaluating it both in zero-shot mode (pre-trained on ASVSPoof-2019; \citet{wang2020asvspoof2019largescalepublic}) and after training with video-level labels from our dataset. BA-TFD and BA-TFD+ (\citet{caiYou2022}) are also evaluated using segmentation proposals treated as frame-level predictions and aggregated by max-voting, both pre-trained on AV-1M and fine-tuned on our dataset.

\begin{table}[t]
\centering
\caption{ Deepfake detection results on the test set of ArEnAV. }
\setlength{\tabcolsep}{2pt}
\scalebox{0.9}{
\begin{tabular}{l|l|l|c|cc|cc|cc}
\toprule[0.4mm]
\rowcolor{mygray} \textbf{Label Access} &\textbf{Pretraining Data}& \textbf{Methods}  & \textbf{Mod.} & \multicolumn{2}{c|}{\textbf{Fullset}} & \multicolumn{2}{c|}{\textbf{Subset V}}  & \multicolumn{2}{c}{\textbf{Subset A}} \\ 
\rowcolor{mygray} \textbf{For Training}  & & &  & AUC & Acc.  & AUC & Acc.  & AUC & Acc. \\ \hline \hline
Zero-Shot &ASVSpoof-19 & XLSR-Mamba & 
 A & 39.19& 52.77 & 52.73  &40.68  & 52.50  & 42.59  \\
- &Internet Scale& Video-LLaMA (7B) & V & 51.48   & 26.29  & 51.47  & 34.21 & 51.43  & 34.18 \\
- &Internet Scale& Video-LLaMA (7B) & AV  & 48.79 & 59.29  & 48.71 &55.37& 48.86 & 55.26\\
- &AV-1M &BA-TFD & AV &61.73  &26.00  &66.42  &34.07  &59.36  & 33.97  \\
- &AV-1M &BA-TFD+ & AV & 60.96 & 25.84  & 64.49  & 34.28  & 59.44 & 33.80  \\
\midrule
Video Level & ArEnAV & XLSR-Mamba & A & 73.00 & 61.00 & 57.47  & 66.16  & 86.33 & 78.00\\
 - &ArEnAV &Meso4 & V & 49.30  & 75.00 & 49.15  & 66.67 & 49.30  & 66.67  \\
 - &ArEnAV &MesoInception4 & V &50.34  & 46.23  & 50.28  & 47.48  & 50.35  & 47.67  \\
 - &ArEnAV & Xception & V & 50.05 & 75.00 & 49.90 & 66.67 & 50.32 & 66.67 \\

 \midrule
 Frame level &ArEnAV &Meso4 & V & 49.55 & 26.60 & 49.60 & 34.40 & 49.53 & 34.36 \\
 - &ArEnAV &MesoInception4 & V & 51.14 & 41.25 & 50.77 & 51.84 & 45.28 & 44.09 \\
 - &ArEnAV & Xception & V & 74.21 & 52.09 & 85.36 & 67.22 & 68.59 & 51.70 \\
 - &AV-1M \& ArEnAV &BA-TFD & AV & 75.91 & 44.31 & 77.64 & 58.29 & 72.21 & 45.21  \\ 
 - &AV-1M \& ArEnAV &BA-TFD+ & AV & 79.97 &27.44  & 84.20  & 36.47  & 72.89 & 34.56  \\
 
\bottomrule[0.1mm]

\end{tabular}}
\vspace{-1em}
\label{tab:classification}
\end{table}

\begin{table*}[b]
  \centering
  \vspace{-2mm}
  \caption{Temporal localization results on ArEnAv, AV-1M and LAV-DF. The low performance on ArEnAV demonstrates the data complexity in CSW settings.}
  \label{tab:cross_temporal_comp}
  \scalebox{0.85}{
    \begin{tabular}{lc||ccc|ccc}
      \toprule[0.4mm]
      \rowcolor{mygray}
      \textbf{Method} & \textbf{Dataset}
        & \textbf{AP@0.5} & \textbf{AP@0.75} & \textbf{AP@0.95}
        & \textbf{AR@50}  & \textbf{AR@20}   & \textbf{AR@10}   \\
      \midrule
      \multirow{3}{*}{BA-TFD}
        & LAV-DF     & 79.15 & 38.57 & 0.24 & 64.18 & 60.89 & 58.51 \\
        & AV-1M & 37.37 & 6.34  & 0.02 & 45.55 & 35.95 & 30.66 \\
        & \textbf{ArEnAV}& \textbf{2.42}  & \textbf{0.55}  & \textbf{0.01} & \textbf{22.30 }& \textbf{3.41}  & \textbf{2.54 } \\
        
      \midrule
      \multirow{3}{*}{BA-TFD+}
        & LAV-DF      & 96.30 & 84.96 & 4.44 & 80.48 & 79.40 & 78.75 \\
        & AV-1M   & 44.42 & 13.64 & 0.03 & 48.86 & 40.37 & 34.67 \\
        & \textbf{ArEnAV} & \textbf{3.74}  &\textbf{ 1.10}  & \textbf{0.04} & \textbf{30.75} & \textbf{4.55 } & \textbf{3.05 } \\
      \bottomrule[0.4mm]
    \end{tabular}}%
  
\end{table*}

\vspace{-1mm}\section{Results and Analysis}
\textbf{Audio-Visual Temporal Deepfake Localization.} The results for temporal localization are shown in Table \ref{tab:temporal_localization}. SOTA methods show significantly lower performance on ArEnAV as compared to other localization datasets (refer to Table \ref{tab:cross_temporal_comp}). BA-TFD and BA-TFD+, 
pretrained on AV-1M, 
show a drop in performance of more than 35\% for AP@0.5 threshold, compared to evaluation on AV-1M. The image-based models, Meso4 and MesoInception4, also provide low performance,
which can be attributed to the use of diffusion-based lip-sync models, which have been overlooked in previous data generation pipelines (\citet{caiAVDeepfake1M2023,caiGlitch2023}). Through this benchmark, we claim that the highly realistic multimodal multilingual code-switched fake content in ArEnAV will open an avenue for further research on temporal multilingual deepfake localization methods.

\textbf{Audio-Visual Deepfake Detection.}
The detection results are in Table \ref{tab:classification}. Image based models, that have access to video-level labels only, perform considerably worse, except XLSR-Mamba, 
which is designed to be trained on video-level labels for audio-deepfake detection. The best performing model is BA-TFD, pretrained on AV-1M 
and then further fine-tuned on our dataset, with AUC Score of 82\% on the full subset. 
We also evaluate models on subsets V and A, as described in the implementation details. The audio-only model, XLSR-Mamba, 
performs better in the Audio-only \emph {subset A}, while the image-only models perform better on \emph{Subset V} for frame-level labels, compared to the \emph{fullset}. XLSR-Mamba performs relatively worst when the audio is code-switched, compared to only Arabic.

\textbf{Cross-Dataset Comparison for Deepfake Localization.} Table \ref{tab:cross_temporal_comp} shows the performance of BA-TFD and BA-TFD+ on LAVDF, AV-1M and ArEnAV datasets. Both models perform significantly worse on ArEnAV, highlighting the poor generalizability in multilingual and code-switching settings. BA-TFD and BA-TFD+ fail to generalize effectively, as the pretrained audio and video encoders struggle with out-of-distribution data encountered in both modalities of ArEnAV.

\vspace{-2mm}
\section{Conclusion}
\vspace{-1mm}
This paper presents ArEnAV, a large multilingual and the first code-switching audio-visual dataset for temporal deepfake localization and detection. The comprehensive benchmark
of the dataset utilizing SOTA deepfake detection and localization methods indicates a significant drop in performance compared to previous monolingual datasets \citet{caiAVDeepfake1M2023,caiGlitch2023}, indicating that the proposed dataset is an important asset for building the next-generation of multilingual deepfake localization methods. As future work, we will evaluate LLM-based detectors after fine-tuning them on the dataset.

\textbf{\textit{Limitations}}\label{section: limitations}. Similar to other deepfake datasets, ArEnAV
exhibits a misbalance in terms of the number of fake and real videos. Due to the limited performance of current SOTA Active-Voice-Detection (Whisper v2) models on Arabic (compared to English), the data generation pipeline can result in a few noisy transcripts. Due to limited instruction following in code-switching scenarios, LLMs might not produce the desired results, as visible in Table \ref{tab:entailment} "Meaning + Translation Scenario". Compared to other subsets, Chat-GPT often fails to follow both instructions, making real and fake transcripts too similar and not always changing their meaning. Also, the dataset is currently limited to two languages only, where we hope to motivate further research in this direction.

\textbf{\textit{Broader Impact.}} ArEnAV’s diverse and realistic English-Arabic fake videos will support the development of more robust audio-visual deepfake detection and localization models, better equipped to handle code-switched speech and real-world multilingual scenarios.

\textbf{\textit{Ethics Statement.}} \label{section: ethics_limitations_broader} We acknowledge that ArEnAV may
raise ethical concerns such as the potential misuse of facial videos
of celebrities, and even the data generation pipeline could have
a potential negative impact. Misuse could include the creation of
deepfake videos or other forms of exploitation. To avoid such issues, we have taken several measures such as distributing the data
with a proper end-user license agreement, where we will impose
certain restrictions on the usage of the data. Furthermore, the user study follows the university IRB guidelines.

\newpage

\bibliographystyle{plainnat} 
\bibliography{reference}

\begin{thebibliography}{50}
\providecommand{\natexlab}[1]{#1}
\providecommand{\url}[1]{\texttt{#1}}
\expandafter\ifx\csname urlstyle\endcsname\relax
  \providecommand{\doi}[1]{doi: #1}\else
  \providecommand{\doi}{doi: \begingroup \urlstyle{rm}\Url}\fi

\bibitem[Afchar et~al.(2018)Afchar, Nozick, Yamagishi, and Echizen]{afcharMesoNet2018}
Darius Afchar, Vincent Nozick, Junichi Yamagishi, and Isao Echizen.
\newblock {MesoNet}: a {Compact} {Facial} {Video} {Forgery} {Detection} {Network}.
\newblock In \emph{2018 {IEEE} {International} {Workshop} on {Information} {Forensics} and {Security} ({WIFS})}, pages 1--7, December 2018.
\newblock ISSN: 2157-4774.

\bibitem[Baevski et~al.(2020)Baevski, Zhou, Mohamed, and Auli]{baevski2020wav2vec20frameworkselfsupervised}
Alexei Baevski, Henry Zhou, Abdelrahman Mohamed, and Michael Auli.
\newblock wav2vec 2.0: A framework for self-supervised learning of speech representations, 2020.
\newblock URL \url{https://arxiv.org/abs/2006.11477}.

\bibitem[Cai et~al.(2022)Cai, Stefanov, Dhall, and Hayat]{caiYou2022}
Zhixi Cai, Kalin Stefanov, Abhinav Dhall, and Munawar Hayat.
\newblock Do {You} {Really} {Mean} {That}? {Content} {Driven} {Audio}-{Visual} {Deepfake} {Dataset} and {Multimodal} {Method} for {Temporal} {Forgery} {Localization}.
\newblock In \emph{2022 {International} {Conference} on {Digital} {Image} {Computing}: {Techniques} and {Applications} ({DICTA})}, pages 1--10, Sydney, Australia, November 2022.

\bibitem[Cai et~al.(2023{\natexlab{a}})Cai, Ghosh, Adatia, Hayat, Dhall, and Stefanov]{caiAVDeepfake1M2023}
Zhixi Cai, Shreya Ghosh, Aman~Pankaj Adatia, Munawar Hayat, Abhinav Dhall, and Kalin Stefanov.
\newblock {AV}-{Deepfake1M}: {A} {Large}-{Scale} {LLM}-{Driven} {Audio}-{Visual} {Deepfake} {Dataset}, November 2023{\natexlab{a}}.
\newblock arXiv:2311.15308 [cs].

\bibitem[Cai et~al.(2023{\natexlab{b}})Cai, Ghosh, Dhall, Gedeon, Stefanov, and Hayat]{caiGlitch2023}
Zhixi Cai, Shreya Ghosh, Abhinav Dhall, Tom Gedeon, Kalin Stefanov, and Munawar Hayat.
\newblock Glitch in the matrix: {A} large scale benchmark for content driven audio–visual forgery detection and localization.
\newblock \emph{Computer Vision and Image Understanding}, 236:\penalty0 103818, November 2023{\natexlab{b}}.
\newblock ISSN 1077-3142.

\bibitem[Cai et~al.(2024)Cai, Dhall, Ghosh, Hayat, Kollias, Stefanov, and Tariq]{cai20241mdeepfakesdetectionchallenge}
Zhixi Cai, Abhinav Dhall, Shreya Ghosh, Munawar Hayat, Dimitrios Kollias, Kalin Stefanov, and Usman Tariq.
\newblock 1m-deepfakes detection challenge, 2024.
\newblock URL \url{https://arxiv.org/abs/2409.06991}.

\bibitem[Casanova et~al.(2023)Casanova, Weber, Shulby, Junior, Gölge, and Ponti]{casanova2023yourttszeroshotmultispeakertts}
Edresson Casanova, Julian Weber, Christopher Shulby, Arnaldo~Candido Junior, Eren Gölge, and Moacir~Antonelli Ponti.
\newblock Yourtts: Towards zero-shot multi-speaker tts and zero-shot voice conversion for everyone, 2023.
\newblock URL \url{https://arxiv.org/abs/2112.02418}.

\bibitem[Casanova et~al.(2024)Casanova, Davis, Gölge, Göknar, Gulea, Hart, Aljafari, Meyer, Morais, Olayemi, and Weber]{casanova2024xttsmassivelymultilingualzeroshot}
Edresson Casanova, Kelly Davis, Eren Gölge, Görkem Göknar, Iulian Gulea, Logan Hart, Aya Aljafari, Joshua Meyer, Reuben Morais, Samuel Olayemi, and Julian Weber.
\newblock Xtts: a massively multilingual zero-shot text-to-speech model, 2024.
\newblock URL \url{https://arxiv.org/abs/2406.04904}.

\bibitem[Chollet(2017)]{cholletXception2017}
Francois Chollet.
\newblock Xception: {Deep} {Learning} {With} {Depthwise} {Separable} {Convolutions}.
\newblock In \emph{Proceedings of the {IEEE} {Conference} on {Computer} {Vision} and {Pattern} {Recognition}}, pages 1251--1258, 2017.

\bibitem[Defossez et~al.(2020)Defossez, Synnaeve, and Adi]{defossez2020real}
Alexandre Defossez, Gabriel Synnaeve, and Yossi Adi.
\newblock Real time speech enhancement in the waveform domain.
\newblock In \emph{Interspeech}, 2020.

\bibitem[Dolhansky et~al.(2020{\natexlab{a}})Dolhansky, Bitton, Pflaum, Lu, Howes, Wang, and Ferrer]{dolhansky2020deepfakedetectionchallengedfdc}
Brian Dolhansky, Joanna Bitton, Ben Pflaum, Jikuo Lu, Russ Howes, Menglin Wang, and Cristian~Canton Ferrer.
\newblock The deepfake detection challenge (dfdc) dataset, 2020{\natexlab{a}}.
\newblock URL \url{https://arxiv.org/abs/2006.07397}.

\bibitem[Dolhansky et~al.(2020{\natexlab{b}})Dolhansky, Bitton, Pflaum, Lu, Howes, Wang, and Ferrer]{dolhanskyDeepFake2020}
Brian Dolhansky, Joanna Bitton, Ben Pflaum, Jikuo Lu, Russ Howes, Menglin Wang, and Cristian~Canton Ferrer.
\newblock The {DeepFake} {Detection} {Challenge} ({DFDC}) {Dataset}, October 2020{\natexlab{b}}.
\newblock arXiv: 2006.07397 [cs].

\bibitem[Face(2025)]{huggingface_speech_df_arena}
Hugging Face.
\newblock Speech df arena - speech-arena-2025.
\newblock \url{https://huggingface.co/spaces/Speech-Arena-2025/Speech-DF-Arena}, 2025.
\newblock Accessed: 2025-05-13.

\bibitem[Frank and Schönherr(2021)]{frank2021wavefakedatasetfacilitate}
Joel Frank and Lea Schönherr.
\newblock Wavefake: A data set to facilitate audio deepfake detection, 2021.
\newblock URL \url{https://arxiv.org/abs/2111.02813}.

\bibitem[Hamed et~al.(2020)Hamed, Vu, and Abdennadher]{hamed-etal-2020-arzen}
Injy Hamed, Ngoc~Thang Vu, and Slim Abdennadher.
\newblock {A}rz{E}n: A speech corpus for code-switched {E}gyptian {A}rabic-{E}nglish.
\newblock In Nicoletta Calzolari, Fr{\'e}d{\'e}ric B{\'e}chet, Philippe Blache, Khalid Choukri, Christopher Cieri, Thierry Declerck, Sara Goggi, Hitoshi Isahara, Bente Maegaard, Joseph Mariani, H{\'e}l{\`e}ne Mazo, Asuncion Moreno, Jan Odijk, and Stelios Piperidis, editors, \emph{Proceedings of the Twelfth Language Resources and Evaluation Conference}, pages 4237--4246, Marseille, France, May 2020. European Language Resources Association.
\newblock ISBN 979-10-95546-34-4.
\newblock URL \url{https://aclanthology.org/2020.lrec-1.523/}.

\bibitem[Hamed et~al.(2024)Hamed, Eryani, Palfreyman, and Habash]{hamed2024zaebuc}
Injy Hamed, Fadhl Eryani, David Palfreyman, and Nizar Habash.
\newblock {ZAEBUC-Spoken}: A multilingual multidialectal {A}rabic-{E}nglish speech corpus.
\newblock In \emph{Proceedings of the 2024 Joint International Conference on Computational Linguistics, Language Resources and Evaluation (LREC-COLING 2024)}, pages 17770--17782, 2024.

\bibitem[Hamed et~al.(2025)Hamed, Sabty, Abdennadher, Vu, Solorio, and Habash]{hamed2025survey}
Injy Hamed, Caroline Sabty, Slim Abdennadher, Ngoc~Thang Vu, Thamar Solorio, and Nizar Habash.
\newblock A survey of code-switched {A}rabic {NLP}: Progress, challenges, and future directions.
\newblock In \emph{Proceedings of the 31st International Conference on Computational Linguistics}, pages 4561--4585, 2025.

\bibitem[He et~al.(2021)He, Gan, Chen, Zhou, Yin, Song, Sheng, Shao, and Liu]{heForgeryNet2021}
Yinan He, Bei Gan, Siyu Chen, Yichun Zhou, Guojun Yin, Luchuan Song, Lu~Sheng, Jing Shao, and Ziwei Liu.
\newblock {ForgeryNet}: {A} {Versatile} {Benchmark} for {Comprehensive} {Forgery} {Analysis}.
\newblock In \emph{Proceedings of the {IEEE}/{CVF} {Conference} on {Computer} {Vision} and {Pattern} {Recognition}}, pages 4360--4369, 2021.

\bibitem[Hou et~al.(2024)Hou, Fu, Chen, Li, Zhang, and Zhao]{hou2024polyglotfakenovelmultilingualmultimodal}
Yang Hou, Haitao Fu, Chuankai Chen, Zida Li, Haoyu Zhang, and Jianjun Zhao.
\newblock Polyglotfake: A novel multilingual and multimodal deepfake dataset, 2024.
\newblock URL \url{https://arxiv.org/abs/2405.08838}.

\bibitem[Jiang et~al.(2020)Jiang, Li, Wu, Qian, and Loy]{jiangDeeperForensics12020}
Liming Jiang, Ren Li, Wayne Wu, Chen Qian, and Chen~Change Loy.
\newblock {DeeperForensics}-1.0: {A} {Large}-{Scale} {Dataset} for {Real}-{World} {Face} {Forgery} {Detection}.
\newblock In \emph{Proceedings of the {IEEE}/{CVF} {Conference} on {Computer} {Vision} and {Pattern} {Recognition}}, pages 2889--2898, 2020.

\bibitem[Khalid et~al.(2022)Khalid, Tariq, Kim, and Woo]{khalid2022fakeavcelebnovelaudiovideomultimodal}
Hasam Khalid, Shahroz Tariq, Minha Kim, and Simon~S. Woo.
\newblock Fakeavceleb: A novel audio-video multimodal deepfake dataset, 2022.
\newblock URL \url{https://arxiv.org/abs/2108.05080}.

\bibitem[Kwon et~al.(2021)Kwon, You, Nam, Park, and Chae]{kwonKoDF2021}
Patrick Kwon, Jaeseong You, Gyuhyeon Nam, Sungwoo Park, and Gyeongsu Chae.
\newblock {KoDF}: {A} {Large}-{Scale} {Korean} {DeepFake} {Detection} {Dataset}.
\newblock In \emph{Proceedings of the {IEEE}/{CVF} {International} {Conference} on {Computer} {Vision}}, pages 10744--10753, 2021.

\bibitem[Li et~al.(2025)Li, Zhang, Xu, Lin, Xie, Feng, Peng, Chen, and Xing]{li2025latentsynctamingaudioconditionedlatent}
Chunyu Li, Chao Zhang, Weikai Xu, Jingyu Lin, Jinghui Xie, Weiguo Feng, Bingyue Peng, Cunjian Chen, and Weiwei Xing.
\newblock Latentsync: Taming audio-conditioned latent diffusion models for lip sync with syncnet supervision, 2025.
\newblock URL \url{https://arxiv.org/abs/2412.09262}.

\bibitem[Li et~al.(2020)Li, Yang, Sun, Qi, and Lyu]{liCelebDF2020}
Yuezun Li, Xin Yang, Pu~Sun, Honggang Qi, and Siwei Lyu.
\newblock Celeb-{DF}: {A} {Large}-{Scale} {Challenging} {Dataset} for {DeepFake} {Forensics}.
\newblock In \emph{Proceedings of the {IEEE}/{CVF} {Conference} on {Computer} {Vision} and {Pattern} {Recognition}}, pages 3207--3216, 2020.

\bibitem[Liu et~al.(2023)Liu, Wang, Sahidullah, Patino, Delgado, Kinnunen, Todisco, Yamagishi, Evans, Nautsch, and Lee]{liuASVspoof2023}
Xuechen Liu, Xin Wang, Md~Sahidullah, Jose Patino, Héctor Delgado, Tomi Kinnunen, Massimiliano Todisco, Junichi Yamagishi, Nicholas Evans, Andreas Nautsch, and Kong~Aik Lee.
\newblock {ASVspoof} 2021: {Towards} {Spoofed} and {Deepfake} {Speech} {Detection} in the {Wild}.
\newblock \emph{IEEE/ACM Transactions on Audio, Speech, and Language Processing}, 31:\penalty0 2507--2522, 2023.
\newblock ISSN 2329-9304.

\bibitem[Marek et~al.(2024)Marek, Kawa, and Syga]{marek2024audiodeepfakedetectionmodels}
Bartłomiej Marek, Piotr Kawa, and Piotr Syga.
\newblock Are audio deepfake detection models polyglots?, 2024.
\newblock URL \url{https://arxiv.org/abs/2412.17924}.

\bibitem[Mubarak et~al.(2021)Mubarak, Hussein, Chowdhury, and Ali]{mubarak2021qasr}
Hamdy Mubarak, Amir Hussein, Shammur~Absar Chowdhury, and Ahmed Ali.
\newblock {QASR}: {QCRI} {Aljazeera} speech resource a large scale annotated {A}rabic speech corpus.
\newblock In \emph{Proceedings of the 59th Annual Meeting of the Association for Computational Linguistics and the 11th International Joint Conference on Natural Language Processing (Volume 1: Long Papers)}, pages 2274--2285, 2021.

\bibitem[Mukhopadhyay et~al.(2023)Mukhopadhyay, Suri, Gadde, and Shrivastava]{mukhopadhyay2023diff2lipaudioconditioneddiffusion}
Soumik Mukhopadhyay, Saksham Suri, Ravi~Teja Gadde, and Abhinav Shrivastava.
\newblock Diff2lip: Audio conditioned diffusion models for lip-synchronization, 2023.
\newblock URL \url{https://arxiv.org/abs/2308.09716}.

\bibitem[Narayan et~al.(2023)Narayan, Agarwal, Thakral, Mittal, Vatsa, and Singh]{narayanDFPlatter2023}
Kartik Narayan, Harsh Agarwal, Kartik Thakral, Surbhi Mittal, Mayank Vatsa, and Richa Singh.
\newblock {DF}-{Platter}: {Multi}-{Face} {Heterogeneous} {Deepfake} {Dataset}.
\newblock In \emph{Proceedings of the {IEEE}/{CVF} {Conference} on {Computer} {Vision} and {Pattern} {Recognition}}, pages 9739--9748, 2023.

\bibitem[Narayan et~al.(2024)Narayan, Djilali, Singh, Bihan, and Hacid]{narayan2024vispermultilingualaudiovisualspeech}
Sanath Narayan, Yasser Abdelaziz~Dahou Djilali, Ankit Singh, Eustache~Le Bihan, and Hakim Hacid.
\newblock Visper: Multilingual audio-visual speech recognition, 2024.
\newblock URL \url{https://arxiv.org/abs/2406.00038}.

\bibitem[Nick and Andrew(2019)]{nickContributing2019}
Dufou Nick and Jigsaw Andrew.
\newblock Contributing {Data} to {Deepfake} {Detection} {Research}, September 2019.

\bibitem[OpenAI(2023)]{openai_tts1}
OpenAI.
\newblock Text-to-speech api: tts-1 model.
\newblock \url{https://platform.openai.com/docs/guides/text-to-speech}, 2023.
\newblock Accessed: 2025-05-13.

\bibitem[OpenAI(2025)]{openai_gpt41_mini}
OpenAI.
\newblock Introducing gpt-4.1 in the api.
\newblock \url{https://openai.com/index/gpt-4-1/}, 2025.
\newblock Accessed: 2025-05-13.

\bibitem[Ott et~al.(2019)Ott, Edunov, Baevski, Fan, Gross, Ng, Grangier, and Auli]{ott-etal-2019-fairseq}
Myle Ott, Sergey Edunov, Alexei Baevski, Angela Fan, Sam Gross, Nathan Ng, David Grangier, and Michael Auli.
\newblock fairseq: A fast, extensible toolkit for sequence modeling.
\newblock In Waleed Ammar, Annie Louis, and Nasrin Mostafazadeh, editors, \emph{Proceedings of the 2019 Conference of the North {A}merican Chapter of the Association for Computational Linguistics (Demonstrations)}, pages 48--53, Minneapolis, Minnesota, June 2019. Association for Computational Linguistics.
\newblock \doi{10.18653/v1/N19-4009}.
\newblock URL \url{https://aclanthology.org/N19-4009/}.

\bibitem[Phukan et~al.(2024)Phukan, Kashyap, Buduru, and Sharma]{phukan2024heterogeneityhomogeneityinvestigatingmultilingual}
Orchid~Chetia Phukan, Gautam~Siddharth Kashyap, Arun~Balaji Buduru, and Rajesh Sharma.
\newblock Heterogeneity over homogeneity: Investigating multilingual speech pre-trained models for detecting audio deepfake, 2024.
\newblock URL \url{https://arxiv.org/abs/2404.00809}.

\bibitem[Qin et~al.(2024)Qin, Zhao, Yu, and Sun]{qin2024openvoiceversatileinstantvoice}
Zengyi Qin, Wenliang Zhao, Xumin Yu, and Xin Sun.
\newblock Openvoice: Versatile instant voice cloning, 2024.
\newblock URL \url{https://arxiv.org/abs/2312.01479}.

\bibitem[Qwen et~al.(2025)Qwen, :, Yang, Yang, Zhang, Hui, Zheng, Yu, Li, Liu, Huang, Wei, Lin, Yang, Tu, Zhang, Yang, Yang, Zhou, Lin, Dang, Lu, Bao, Yang, Yu, Li, Xue, Zhang, Zhu, Men, Lin, Li, Tang, Xia, Ren, Ren, Fan, Su, Zhang, Wan, Liu, Cui, Zhang, and Qiu]{qwen2025qwen25technicalreport}
Qwen, :, An~Yang, Baosong Yang, Beichen Zhang, Binyuan Hui, Bo~Zheng, Bowen Yu, Chengyuan Li, Dayiheng Liu, Fei Huang, Haoran Wei, Huan Lin, Jian Yang, Jianhong Tu, Jianwei Zhang, Jianxin Yang, Jiaxi Yang, Jingren Zhou, Junyang Lin, Kai Dang, Keming Lu, Keqin Bao, Kexin Yang, Le~Yu, Mei Li, Mingfeng Xue, Pei Zhang, Qin Zhu, Rui Men, Runji Lin, Tianhao Li, Tianyi Tang, Tingyu Xia, Xingzhang Ren, Xuancheng Ren, Yang Fan, Yang Su, Yichang Zhang, Yu~Wan, Yuqiong Liu, Zeyu Cui, Zhenru Zhang, and Zihan Qiu.
\newblock Qwen2.5 technical report, 2025.
\newblock URL \url{https://arxiv.org/abs/2412.15115}.

\bibitem[Radford et~al.(2022{\natexlab{a}})Radford, Kim, Xu, Brockman, McLeavey, and Sutskever]{radford2022robustspeechrecognitionlargescale}
Alec Radford, Jong~Wook Kim, Tao Xu, Greg Brockman, Christine McLeavey, and Ilya Sutskever.
\newblock Robust speech recognition via large-scale weak supervision, 2022{\natexlab{a}}.
\newblock URL \url{https://arxiv.org/abs/2212.04356}.

\bibitem[Radford et~al.(2022{\natexlab{b}})Radford, Kim, Xu, Brockman, McLeavey, and Sutskever]{radford2022whisper}
Alec Radford, Jong~Wook Kim, Tao Xu, Greg Brockman, Christine McLeavey, and Ilya Sutskever.
\newblock Robust speech recognition via large-scale weak supervision, 2022{\natexlab{b}}.
\newblock URL \url{https://arxiv.org/abs/2212.04356}.

\bibitem[Rossler et~al.(2019)Rossler, Cozzolino, Verdoliva, Riess, Thies, and Niessner]{rosslerFaceForensics2019}
Andreas Rossler, Davide Cozzolino, Luisa Verdoliva, Christian Riess, Justus Thies, and Matthias Niessner.
\newblock {FaceForensics}++: {Learning} to {Detect} {Manipulated} {Facial} {Images}.
\newblock In \emph{Proceedings of the {IEEE}/{CVF} {International} {Conference} on {Computer} {Vision}}, pages 1--11, 2019.

\bibitem[Rössler et~al.(2019)Rössler, Cozzolino, Verdoliva, Riess, Thies, and Nießner]{rössler2019faceforensicslearningdetectmanipulated}
Andreas Rössler, Davide Cozzolino, Luisa Verdoliva, Christian Riess, Justus Thies, and Matthias Nießner.
\newblock Faceforensics++: Learning to detect manipulated facial images, 2019.
\newblock URL \url{https://arxiv.org/abs/1901.08971}.

\bibitem[SDAIANCAI(2025)]{sdaian_cai_code_switching_dataset}
SDAIANCAI.
\newblock Ar-en code-switching textual dataset.
\newblock \url{https://huggingface.co/datasets/SDAIANCAI/Ar-En-Code-Switching-Textual-Dataset}, 2025.
\newblock Accessed: 2025-05-13.

\bibitem[Sengupta et~al.(2023)Sengupta, Sahu, Jia, Katipomu, Li, Koto, Marshall, Gosal, Liu, Chen, Afzal, Kamboj, Pandit, Pal, Pradhan, Mujahid, Baali, Han, Bsharat, Aji, Shen, Liu, Vassilieva, Hestness, Hock, Feldman, Lee, Jackson, Ren, Nakov, Baldwin, and Xing]{sengupta2023jaisjaischatarabiccentricfoundation}
Neha Sengupta, Sunil~Kumar Sahu, Bokang Jia, Satheesh Katipomu, Haonan Li, Fajri Koto, William Marshall, Gurpreet Gosal, Cynthia Liu, Zhiming Chen, Osama~Mohammed Afzal, Samta Kamboj, Onkar Pandit, Rahul Pal, Lalit Pradhan, Zain~Muhammad Mujahid, Massa Baali, Xudong Han, Sondos~Mahmoud Bsharat, Alham~Fikri Aji, Zhiqiang Shen, Zhengzhong Liu, Natalia Vassilieva, Joel Hestness, Andy Hock, Andrew Feldman, Jonathan Lee, Andrew Jackson, Hector~Xuguang Ren, Preslav Nakov, Timothy Baldwin, and Eric Xing.
\newblock Jais and jais-chat: Arabic-centric foundation and instruction-tuned open generative large language models, 2023.
\newblock URL \url{https://arxiv.org/abs/2308.16149}.

\bibitem[Thakral et~al.(2025)Thakral, Ranjan, Singh, Jain, Singh, and Vatsa]{thakral2025illusion}
Kartik Thakral, Rishabh Ranjan, Akanksha Singh, Akshat Jain, Richa Singh, and Mayank Vatsa.
\newblock {ILLUSION}: Unveiling truth with a comprehensive multi-modal, multi-lingual deepfake dataset.
\newblock In \emph{The Thirteenth International Conference on Learning Representations}, 2025.
\newblock URL \url{https://openreview.net/forum?id=qnlG3zPQUy}.

\bibitem[Thies et~al.(2020)Thies, Zollhöfer, Stamminger, Theobalt, and Nießner]{thies2020face2facerealtimefacecapture}
Justus Thies, Michael Zollhöfer, Marc Stamminger, Christian Theobalt, and Matthias Nießner.
\newblock Face2face: Real-time face capture and reenactment of rgb videos, 2020.
\newblock URL \url{https://arxiv.org/abs/2007.14808}.

\bibitem[Todisco et~al.(2019)Todisco, Wang, Vestman, Sahidullah, Delgado, Nautsch, Yamagishi, Evans, Kinnunen, and Lee]{todiscoASVspoof2019}
Massimiliano Todisco, Xin Wang, Ville Vestman, Md~Sahidullah, Hector Delgado, Andreas Nautsch, Junichi Yamagishi, Nicholas Evans, Tomi Kinnunen, and Kong~Aik Lee.
\newblock {ASVspoof} 2019: {Future} {Horizons} in {Spoofed} and {Fake} {Audio} {Detection}, April 2019.
\newblock arXiv:1904.05441 [cs, eess].

\bibitem[Wang et~al.(2020)Wang, Yamagishi, Todisco, Delgado, Nautsch, Evans, Sahidullah, Vestman, Kinnunen, Lee, Juvela, Alku, Peng, Hwang, Tsao, Wang, Maguer, Becker, Henderson, Clark, Zhang, Wang, Jia, Onuma, Mushika, Kaneda, Jiang, Liu, Wu, Huang, Toda, Tanaka, Kameoka, Steiner, Matrouf, Bonastre, Govender, Ronanki, Zhang, and Ling]{wang2020asvspoof2019largescalepublic}
Xin Wang, Junichi Yamagishi, Massimiliano Todisco, Hector Delgado, Andreas Nautsch, Nicholas Evans, Md~Sahidullah, Ville Vestman, Tomi Kinnunen, Kong~Aik Lee, Lauri Juvela, Paavo Alku, Yu-Huai Peng, Hsin-Te Hwang, Yu~Tsao, Hsin-Min Wang, Sebastien~Le Maguer, Markus Becker, Fergus Henderson, Rob Clark, Yu~Zhang, Quan Wang, Ye~Jia, Kai Onuma, Koji Mushika, Takashi Kaneda, Yuan Jiang, Li-Juan Liu, Yi-Chiao Wu, Wen-Chin Huang, Tomoki Toda, Kou Tanaka, Hirokazu Kameoka, Ingmar Steiner, Driss Matrouf, Jean-Francois Bonastre, Avashna Govender, Srikanth Ronanki, Jing-Xuan Zhang, and Zhen-Hua Ling.
\newblock Asvspoof 2019: A large-scale public database of synthesized, converted and replayed speech, 2020.
\newblock URL \url{https://arxiv.org/abs/1911.01601}.

\bibitem[Wang et~al.(2017)Wang, Skerry-Ryan, Stanton, Wu, Weiss, Jaitly, Yang, Xiao, Chen, Bengio, Le, Agiomyrgiannakis, Clark, and Saurous]{wang2017tacotronendtoendspeechsynthesis}
Yuxuan Wang, RJ~Skerry-Ryan, Daisy Stanton, Yonghui Wu, Ron~J. Weiss, Navdeep Jaitly, Zongheng Yang, Ying Xiao, Zhifeng Chen, Samy Bengio, Quoc Le, Yannis Agiomyrgiannakis, Rob Clark, and Rif~A. Saurous.
\newblock Tacotron: Towards end-to-end speech synthesis, 2017.
\newblock URL \url{https://arxiv.org/abs/1703.10135}.

\bibitem[Xiao and Das(2025)]{xiao2025xlsrmambadualcolumnbidirectionalstate}
Yang Xiao and Rohan~Kumar Das.
\newblock Xlsr-mamba: A dual-column bidirectional state space model for spoofing attack detection, 2025.
\newblock URL \url{https://arxiv.org/abs/2411.10027}.

\bibitem[Zhang et~al.(2023)Zhang, Li, and Bing]{zhangVideoLLaMA2023}
Hang Zhang, Xin Li, and Lidong Bing.
\newblock Video-{LLaMA}: {An} {Instruction}-tuned {Audio}-{Visual} {Language} {Model} for {Video} {Understanding}, June 2023.
\newblock arXiv:2306.02858 [cs, eess].

\end{thebibliography}

\newpage

\appendix

\section{Real Perturbations}

\begin{table}[htbp]
\centering
\caption{List of video and audio perturbation types with descriptions.}
\small
\renewcommand{\arraystretch}{1.3}
\begin{tabular}{|c|c|l|}
\hline
\textbf{Category} & \textbf{Perturbation Type} & \textbf{Description} \\
\hline
\multirow{15}{*}{\textbf{Video Perturbations}} & Gaussian Blur & Applies Gaussian smoothing to simulate out-of-focus capture. \\
 & Salt and Pepper Noise & Random white and black pixel noise, mimicking sensor errors. \\
 & Low Bitrate Compression & Blocky, artifact-heavy images due to compression. \\
 & Gaussian Noise & Electronic sensor noise typical in low-light conditions. \\
 & Poisson Noise (Shot Noise) & Noise from photon-limited imaging environments. \\
 & Speckle Noise & Multiplicative noise creating granular interference effects. \\
 & Color Quantization & Banding effects from limited color palettes. \\
 & Random Brightness & Simulates variations in exposure and lighting. \\
 & Motion Blur & Imitates camera or object motion during capture. \\
 & Rolling Shutter & Distortion effects due to CMOS sensor movements. \\
 & Camera Shake & Minor frame shifts from handheld camera vibrations. \\
 & Lens Distortion & Optical distortions like barrel or pincushion effects. \\
 & Vignetting & Darkening of image edges typical of certain lenses. \\
 & Exposure Variation & Adjusts brightness and contrast, simulating exposure issues. \\
 & Chromatic Aberration & Color channel shifts causing fringing effects. \\
\hline
\multirow{11}{*}{\textbf{Audio Perturbations}} & Compression Artifacts & Quality loss from low bitrate compression. \\
 & Pitch/Loudness Distortion & Gain or frequency alterations simulating recording issues. \\
 & White Noise & Constant background electronic interference noise. \\
 & Time Stretch & Audio speed adjustments without pitch change. \\
 & Reverberation & Echo and reverb modeling room acoustics. \\
 & Ambient Noise & Background environmental sounds added. \\
 & Clipping & Distortion from exceeding audio amplitude limits. \\
 & Frequency Filter & Filtering effects simulating transmission equipment variations. \\
 & Doppler Effect & Pitch modulation due to relative motion. \\
 & Interference & Static-like bursts mimicking external disturbances. \\
 & Room Impulse Response & Complex echo patterns modeling specific environments. \\
\hline
\end{tabular}

\label{tab:perturbations}
\end{table}

\label{lab:Appendix}
\newpage
\section{Augmentation Examples}
\label{sec:appendix_aug_examples}
In Table \ref{tab:augmentation_examples}, we provide examples of augmentations achieved through the manipulation rules previously outlined in Section \ref{sec:transcript_manipulation}.
\begin{table}[h]
\centering
\caption{Examples of augmentations achieved through the different transcript manipulation rules, showing the original (orig) and augmented (aug) transcriptions.}
\label{tab:augmentation_examples}
 \includegraphics[width=\textwidth]{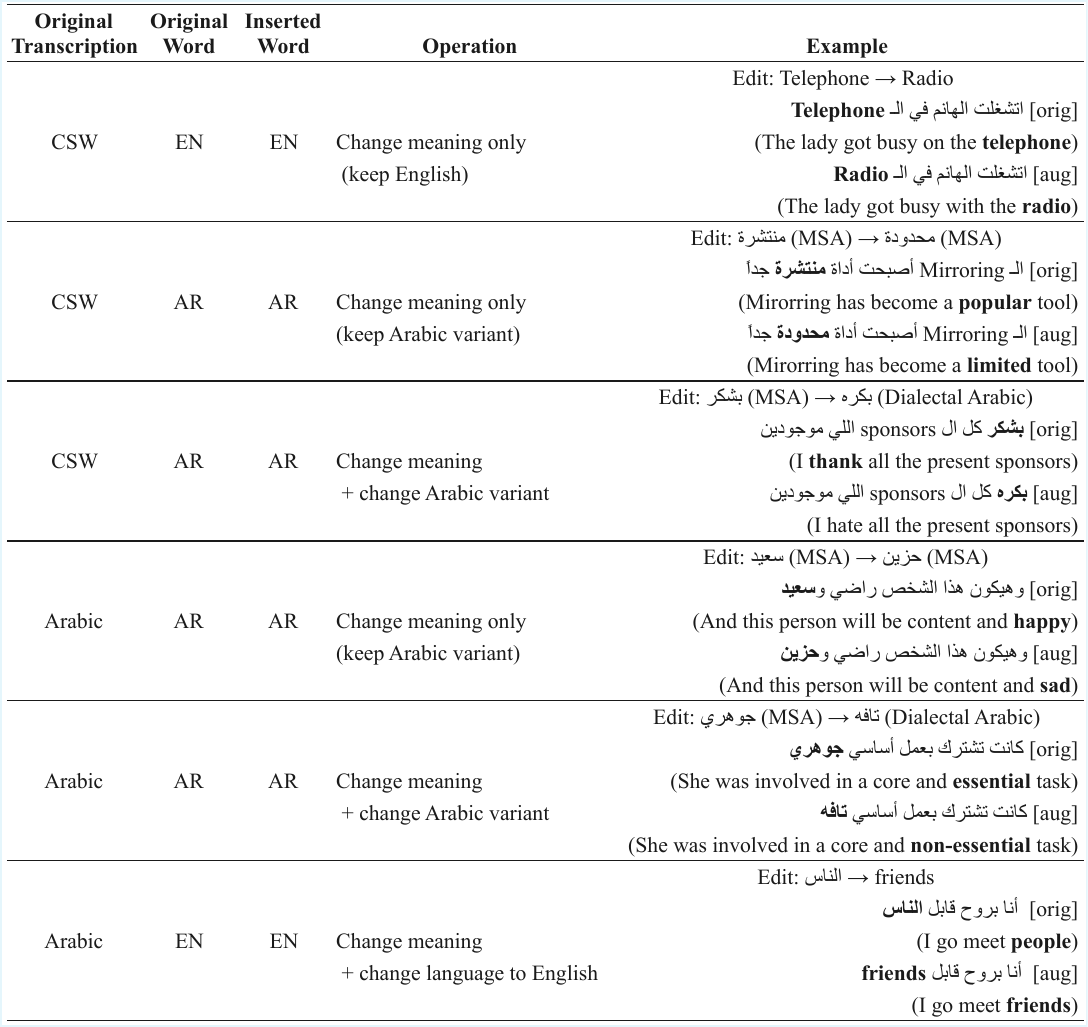}
\end{table}
\newpage
\section{Prompt for Text Perturbation}
\label{sec:prompt}
\begin{figure}[htbp]
    \centering
    \begin{tcolorbox}[colback=gray!5!white, colframe=gray!75!black, title=Prompt for Fake Transcript Generation.]
\small 
\begin{verbatim}
###SYSTEM MESSAGE###
You are a controlled text-perturbation bot.
Here is the transcript of an audio. 
Please use the provided operations to modify
the transcript to change its sentiment. 
The operation can be one of `delete`,
`insert` and `replace`. 
Please priority modify adjectives and adverbs.
-------------------CHANGE-MODES------------------
• meaning_only
      - Change the *meaning* of one word.
      - Keep the same language/script and dialect.
• dialect_only
      - Swap a word for a dialectal equivalent of *identical meaning*.
      - Example: <syArT> → <`rbyT> (Gulf dialect, same meaning).
• dialect_plus_meaning
      - Change *both* dialect *and* meaning in a single word.
      - Example: <jmyl> (msa, 'nice') → <wH$> (Egyptian, 'awful').
• meaning_plus_translation
      - In Arabic-only sentences, pick a word that 
      is **commonly code-switched
to English** in everyday speech (e.g., <mwbayl>, <syArT>, <Antrnt>).
      - Translate that word to English and change the 
      meaning simultaneously. 
        Example: <syArT> ('car') → bike.
-------------------CSW MULTI-OP LOGIC-------------------
If language == 'csw':
  num = 1  → edit exactly one token matching target_token_script.
  num = 2  → edit 1 English + 1 Arabic token.
  num = 3  → edit 1 English + 2 Arabic tokens.
 -------------------OTHER RULES-------------------
• Only modify tokens that are *commonly code-switched* in real speech
  (brand names, technology, everyday nouns, etc.).
• Each operation targets ONE word (delete / insert / replace).
• Number of operations for INSERT, DELETE and REPLACE 
should be equal across
the data.
• If sentiment can be changed with INSERT or DELETE, 
prefer it over REPLACE.
• When dialect shifts, include original_dialect and new_dialect.
• Never alter tense or add restricted content.
• Return **only** a JSON object that matches the schema.
\end{verbatim}
    \end{tcolorbox}
    \caption{System prompt for text-perturbation bot}
    \label{fig:system_prompt}
\end{figure}

\end{document}